\documentclass[10pt,twocolumn,letterpaper]{article}

\usepackage[pagenumbers]{cvpr} 
\usepackage[accsupp]{axessibility}

\usepackage{enumitem}
\usepackage{indentfirst}
\usepackage{tabularx}
\usepackage{booktabs}
\usepackage{multirow}
\usepackage{color}
\usepackage[normalem]{ulem}
\useunder{\uline}{\ul}{}

\usepackage[dvipsnames]{xcolor}

\definecolor{cvprblue}{rgb}{0.21,0.49,0.74}
\usepackage[pagebackref,breaklinks,colorlinks,citecolor=cvprblue]{hyperref}

\def\paperID{11603} 
\def\confName{CVPR}
\def\confYear{2024}

\title{Incremental Residual Concept Bottleneck Models}

\author{Chenming Shang$^{12}$\footnotemark[1] \quad Shiji Zhou$^1$ \quad Hengyuan Zhang$^1$ \quad Xinzhe Ni$^1$ \quad Yujiu Yang$^1$\footnotemark[2] \quad Yuwang Wang$^{12}$\footnotemark[2]\\
$^1$Tsinghua University \quad $^2$Shanghai AI Laboratory\\
{\tt\small \{scm22@mails.,zhoushiji@mail.,zhang-hy22@mails.\}tsinghua.edu.cn}\\
{\tt\small \{nxz22@mails.,yang.yujiu@sz.,wang-yuwang@mail.\}tsinghua.edu.cn}
}

\begin{document}

\maketitle
\renewcommand{\thefootnote}{\fnsymbol{footnote}}
\footnotetext[1]{Work done during an internship at Shanghai AI Laboratory.}
\footnotetext[2]{Corresponding author.}

\maketitle
\renewcommand{\thefootnote}{\arabic{footnote}}
\begin{abstract}

Concept Bottleneck Models (CBMs) map the black-box visual representations extracted by deep neural networks onto a set of interpretable concepts and use the concepts to make predictions, enhancing the transparency of the decision-making process. Multimodal pre-trained models can match visual representations with textual concept embeddings, allowing for obtaining the interpretable concept bottleneck without the expertise concept annotations. Recent research has focused on the concept bank establishment and the high-quality concept selection. However, it is challenging to construct a comprehensive concept bank through humans or large language models, which severely limits the performance of CBMs. In this work, we propose the Incremental \textbf{Res}idual \textbf{C}oncept \textbf{B}ottleneck \textbf{M}odel (Res-CBM) to address the challenge of concept completeness. Specifically, the residual concept bottleneck model employs a set of optimizable vectors to complete missing concepts, then the incremental concept discovery module converts the complemented vectors with unclear meanings into potential concepts in the candidate concept bank. Our approach can be applied to any user-defined concept bank, as a post-hoc processing method to enhance the performance of any CBMs. Furthermore, to measure the descriptive efficiency of CBMs, the \textbf{C}oncept \textbf{U}tilization \textbf{E}fficiency (CUE) metric is proposed. Experiments show that the Res-CBM outperforms the current state-of-the-art methods in terms of both accuracy and efficiency and achieves comparable performance to black-box models across multiple datasets. \footnote{The code is available at: \url{https://github.com/HelloSCM/Res-CBM}.}

\end{abstract}    
\section{Introduction}
\label{sec:introduction}

\begin{figure}[t]
    \centering
    \includegraphics[width=\linewidth]{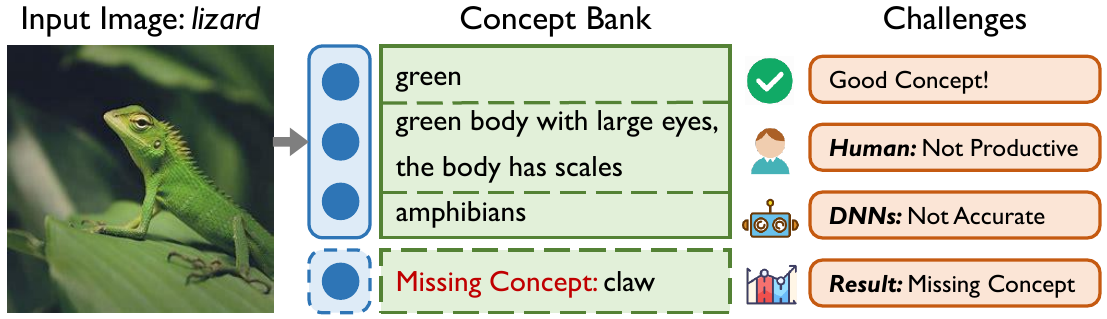}
    \caption{Challenges in CBMs. When a concept contains many atomic attributes, making it too complex for human to comprehend. When a concept is too high-level, DNNs may struggle to provide accurate prediction. Additionally, there is a risk of human-designed concept bank missing some important concepts.}
    \label{fig:challenges}
    \vspace{-1mm}
\end{figure}

Deep neural networks (DNNs) \cite{lecun2015deep,samek2021explaining} have achieved unprecedented success in a wide range of machine learning tasks, including computer vision \cite{han2022survey,li2021survey}, natural language processing \cite{zhang2023assisting,guo2024connecting}, and speech recognition \cite{radford2023robust,malik2021automatic}. However, due to their complex and deep structures, they are often regarded as black-box models \cite{castelvecchi2016can,shwartz2017opening}, which are too difficult to understand and interpret. In fields that demand high levels of trustworthiness, such as medicine \cite{razzak2018deep}, healthcare \cite{9428234}, education \cite{zhang2024questioncentric,zhang2024improving}, and finance \cite{ozbayoglu2020deep}, model interpretability has become increasingly crucial. It indicates whether we can trust the DNNs' decisions, and how we can rectify the errors when the DNNs make mistakes. Making deep learning models more interpretable is a significant yet challenging research topic.

A promising approach for achieving interpretability in deep learning is through concept-based models \cite{schwalbe2022concept}, which leverage human-generated high-level concepts to explain the black-box features of DNNs. Among these approaches, Concept Bottleneck Models (CBMs) \cite{koh2020concept} map the visual representations to a set of concept values that are understandable to humans. These interpretable concepts are then used to make the final decision by a linear function, greatly enhancing our understanding of the decision-making process. Of greater significance, humans can repair models' faults by directly editing or intervening on the concept bottleneck \cite{shin2023closer,bahadori2020debiasing}.

Due to the considerable cost of the fine-grained and precise annotation for each concept in CBMs, multimodal pre-trained model-based CBMs \cite{yuksekgonul2022post} have recently emerged as a research hotspot. CLIP \cite{radford2021learning} demonstrates the capacity to establish correspondences between visual information and textual descriptions. Accordingly, we can utilize the CLIP text encoder to encode human-understood concepts into concept embeddings and use the visual encoder to encode images into visual representations. By projecting visual representations to each concept embedding, we can obtain the concept bottleneck. The recent innovations have primarily concentrated on comprehensive concept bank establishment and efficient high-quality concept selection. However, these methods face three critical challenges as shown in Fig.\ref{fig:challenges}:

\begin{itemize}
  \item \textbf{Purity:} The essence of concepts is to abstract complex information into a combination of simple foundational elements, which can generalize to unseen data through the infinite combinations of finite concepts, offering an efficient way to describe the world. Long and complex descriptions like \textit{small, black insect with six legs} or \textit{long, thin tool with a wooden handle}, can be composed entirely of combinations of simple concepts such as \textit{small}, \textit{black}, \textit{insect}, \textit{leg}, etc., which may increase the productivity of concept utilization. Since these complicated concepts are excessively detailed, they may not be shared across different classes, which will limit their ability to capture shared knowledge between different categories.
  
  \item \textbf{Precision:} Each concept should be clearly understandable for both humans and multimodal pre-trained models when annotating concepts. Specifically, concepts should not be markedly more difficult than classes. For example, high-level concepts like \textit{amphibians} or \textit{perennial}, are relatively harder for humans to understand and exhibit weaker interpretability. Moreover, we cannot ascertain whether CLIP possesses the capability to understand these high-level concepts, which can potentially lead to errors in concept annotation.
  
  \item \textbf{Completeness:} Concept bank should strive to include as much relevant visual information that is beneficial for downstream classification tasks as possible. It is essential to avoid losing too much information in the process. However, it is challenging to establish a comprehensive concept bank directly \cite{yeh2020completeness}. The exhaustive capability of language descriptions is limited, making it difficult to accomplish a fully thorough concept bank through manual design and selection.
\end{itemize}

In this work, we propose the Incremental \textbf{Res}idual \textbf{C}oncept \textbf{B}ottleneck \textbf{M}odel (Res-CBM) to address the three challenges above. First, we establish a simple and pure base concept bank and candidate concept bank to ensure that the CLIP can accurately recognize each concept. Next, we initialize a set of optimizable vectors as filling concepts and optimize this set of filling concepts through the residual concept bottleneck model, to make up for the inadequacy in the base concept bank. Finally, the incremental concept discovery module is utilized to translate these complemented vectors with unclear meanings into potential concepts in the candidate concept bank, improving the performance while preserving the interpretability of the CBM. Our main contributions can be summarized as follows:

\begin{enumerate}
    \item We systematically reviewed the research on the CLIP-based CBMs, clarifying the importance of the concepts' purity, precision, and completeness.
    
    \item We proposed the residual concept bottleneck model to make up for the insufficiency in the base concept bank and converted these complemented unknown vectors into potential concepts in the candidate concept bank through the incremental concept discovery module. It should be emphasized that our approach can be applied to any user-defined concept bank, as a post-hoc processing method to enhance the performance of any CBMs.
    
    \item To measure the descriptive efficiency of CBMs, we introduced the metric of the concept utilization efficiency. We performed extensive experiments to validate our methodology and elaborate ablation studies to illustrate the effectiveness of each component.
\end{enumerate}
\section{Related Work }
\label{sec:related work}

Deep neural networks (DNNs) can be explained through pixels \cite{selvaraju2017grad,chattopadhay2018grad}, samples \cite{hammoudeh2022training,ilyas2022datamodels}, weights \cite{wortsman2020supermasks}, individual neurons \cite{ghorbani2020neuron,blalock2020state}, subnetworks \cite{amer2019review,frankle2018lottery}, representations \cite{bengio2013representation,hendricks2016generating,hendricks2018grounding}, etc \cite{rauker2023toward}. Concept-based models \cite{losch2019interpretability,kazhdan2020now,wang2020chain,wang2023learning,shang2024understanding} are representation-level interpretable approaches, whose core idea is to map representations with unknown meanings extracted by the black-box DNNs to a set of concepts that can be comprehended by humans \cite{margeloiu2021concept,espinosa2022concept,crabbe2022concept,marconato2022glancenets}. Concept-based models can be broadly categorized into Concept Bottleneck Models (CBMs) \cite{koh2020concept}, which describe concepts as independent values annotated by experts, and Concept Activation Vectors (CAVs) \cite{kim2018interpretability}, which represent concepts as normal vectors of the decision boundaries that distinguish positive and negative samples of a concept. CBMs involve the datasets labeled with fine-grained and precise attributes by expert knowledge, such as CUB-200 \cite{he2019fine}, OAI \cite{eckstein2012recent}, and LAD \cite{zhao2019large}, whereas CAVs require learning the concept activation vectors on additional probe datasets, and the directions of the concept activation vectors are strictly correlated with the probe datasets \cite{ramaswamy2022overlooked}. These limitations severely restrict the development of concept-based models.

\begin{figure*}[t]
    \centering
    \includegraphics[width=0.8\linewidth]{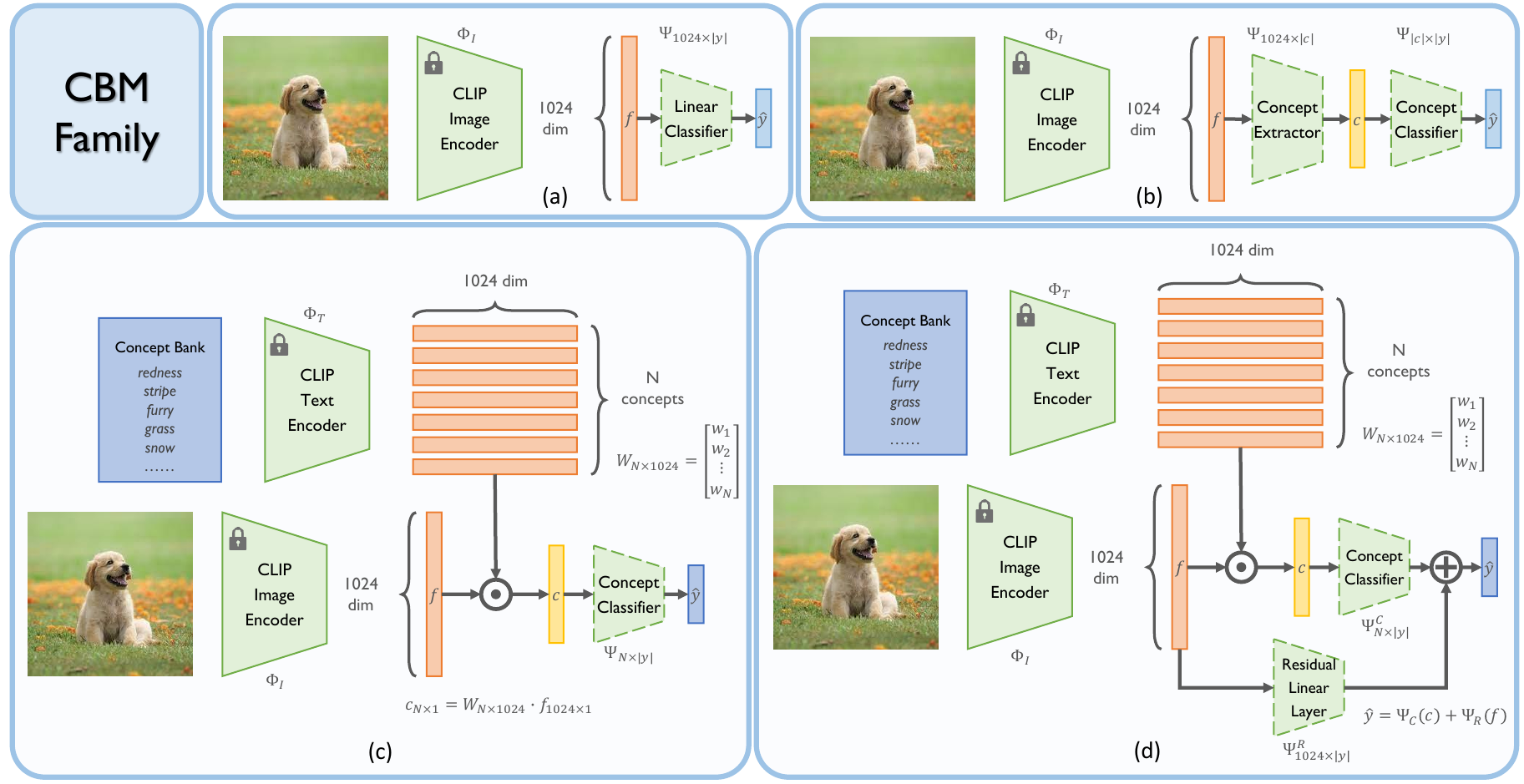}
    \caption{Different CBMs' structures. (a): CLIP linear probing \cite{radford2021learning}. (b): Original Concept Bottleneck Model (CBM) \cite{koh2020concept}. (c) Post-hoc Concept Bottleneck Model (PCBM) \cite{yang2023language,oikarinen2023label,yuksekgonul2022post}. (d): Hybrid Post-hoc Concept Bottleneck Model (PCBM-h) \cite{yuksekgonul2022post}.}
    \label{fig:cbm}
\end{figure*}

Recently, there has been a rapid development of multimodal pre-trained models \cite{han2023survey}, among which CLIP does an excellent job of relating visual information to text through contrastive learning, giving researchers confidence for unsupervised concept labeling with CLIP. PCBM \cite{yuksekgonul2022post} generates a concept bank via ConceptNet \cite{speer2017conceptnet} and calculates the projection distance between the CLIP textual concept embeddings and the CLIP visual representations of images, analogous to the CAVs, to obtain the concept bottleneck. PCBM-h \cite{yuksekgonul2022post} introduces an uninterpretable residual linear layer to compensate for incomplete concept extraction by fitting the difference between the PCBM's result and the ground truth through the original visual representations of CLIP. Label-free CBM \cite{oikarinen2023label} only leverages CLIP to concept annotation, maximizing the similarity between the concept bottleneck of the features extracted by arbitrary visual backbone and the concept annotations labeled by CLIP. LaBo \cite{yang2023language} generates a set of candidate concepts through large language models and designs a submodular optimization block to select the concepts.

\section{Method}
\label{sec:method}


\subsection{Problem Formulation}
\label{subsec:problem definition}
Consider a dataset of image-label pairs $\mathcal{D} = \{(x, y)\}$, where $x \in \mathcal{X}$ is the image and $y \in \mathcal{Y}$ is the label. We have $N$ human-selected concepts to describe the essential information of these images, which can be denoted as discrete tokens $\mathcal{E} = \{e_1,e_2,...,e_N\}$. Multimodal pre-trained alignment model (e.g., CLIP \cite{radford2021learning}) has an image encoder $\Phi_I:\mathcal{X} \rightarrow \mathbb{R}^d$ and a text encoder $\Phi_T$, which can map images and text into a shared $d$-dimensional feature space respectively. We encode the discrete tokens with the CLIP text encoder then perform $L_2$ normalization\footnote{The symbol of $L_2$ normalization is omitted for writing convenience.} to obtain the concept embeddings $\{w_1,w_2,...,w_N | w_i = \Phi_T(e_i), i=1,2,...,N\}$ with the length of $1$ and the dimension of $d$. We concatenate these concept embeddings to a concept projection matrix $W_{N \times d}:\mathbb{R}^d \rightarrow \mathbb{R}^N$ in arbitrary order, also identified as a concept bank.

Correspondingly, we utilize the CLIP image encoder to get the visual representations $f = \Phi_I(x)$. Considering that CLIP has aligned the images with the textual data when pre-training, the visual representations share the feature space with any concept embedding, and the projection length $\Vert f \Vert_2 \cdot \cos\langle w_i,f \rangle$ can reflect the presence of a particular concept in the image. Since undergoing $L_2$ normalization on $w_i$, the concept probability can be directly formulated in terms of the dot product $c=W \cdot f$.

Fig.\ref{fig:cbm} illustrates the structure of common CBMs. For \textbf{CLIP Linear Probing} \cite{kumar2022fine}, an uninterpretable linear classifier $\Psi:\mathbb{R}^d \rightarrow \mathcal{Y}$ is trained directly from visual representations:
\begin{equation}
    \hat{y} = \Psi(f) = \Psi(\Phi_I(x))
\end{equation}

\textbf{Original CBM} \cite{koh2020concept} first extracts concepts using a concept extractor $\Psi_{\mathrm{cpt}}:\mathbb{R}^d \rightarrow \mathbb{R}^N$ and then linearly unites the extracted concepts for label prediction by concept classifier $\Psi_{\mathrm{cls}}:\mathbb{R}^N \rightarrow \mathcal{Y}$.
\begin{equation}
    \hat{y} = \Psi_{\mathrm{cls}}(\Psi_{\mathrm{cpt}}(f))
    \label{eq:cbm}
\end{equation}

\textbf{Post-hoc CBM} \cite{yuksekgonul2022post} stands for a category of CLIP-based CBMs, including Label-free CBM \cite{oikarinen2023label}, LaBo \cite{yang2023language}, etc. They attain concept bottleneck by projecting visual representations $f$ directly to a concept bank $W$, rather than by learning a concept extractor. After getting concepts, only a linear concept classifier $\Psi_c:\mathbb{R}^N \rightarrow \mathcal{Y}$ needs to be learned.
\begin{equation}
    \hat{y} = \Psi_c(c) = \Psi_c(W \cdot f)
\end{equation}

\textbf{PCBM-h} \cite{yuksekgonul2022post} adds a linear layer $\Psi_r:\mathbb{R}^d \rightarrow \mathcal{Y}$ between the original visual representation and the labels on the basis of PCBM, which is used to predict the difference between the PCBM result and the ground truth.
\begin{equation}
    \hat{y} = \Psi_c(c) + \Psi_r(f) = \Psi_c(W \cdot f) + \Psi_r(f)
\end{equation}


\subsection{Concept Bank Establishment}
\label{subsec:concept bank}

\textbf{Candidate Concept Bank.} To ensure precision, we need to make sure that CLIP is capable of recognizing each concept, and one straightforward way is to construct the concept bank from the CLIP pre-trained dataset. For purity, it is desirable for each concept to be at the atom level rather than the compound level, thus maintaining the clarity and specificity. We find that the scene graph \cite{krishna2017visual} consisting of visual concepts to be an appropriate foundation, in which an atom is defined as a singular visual concept, corresponding to a single scene graph node. Atoms are subtyped into \textit{objects}, \textit{relationships}, and \textit{attributes}. We pick the nouns and adjectives in them as the candidate concept bank $\mathcal{E}^\ast$.

\textbf{Base Concept Bank.} Although our method can recover missing concepts from a diverse base concept bank, a high-quality base concept bank can remarkably improve the efficiency of our approach. A base concept bank is expected to consist of two parts. Firstly, it should include general concepts that are applicable across various scenarios. We select the pre-defined concept library proposed by \cite{kim2018interpretability}, which has been widely used in various tasks \cite{abid2022meaningfully,wu2023discover}, as our general concept bank $\mathcal{E}_\mathrm{ge}$. Secondly, it is supposed to include concepts relevant to the specific classification task. Following the PCBM \cite{yuksekgonul2022post}, we collect all concepts from ConceptNet \cite{speer2017conceptnet} that have relations of \textit{hasA}, \textit{isA}, \textit{partOf}, \textit{HasProperty}, and \textit{MadeOf} with the classes in each classification task to build the associated concept bank $\mathcal{E}_\mathrm{as}$. Both concept banks should be included in the candidate concept bank, from which we can derive the base concept bank:
\begin{equation}
    \mathcal{E}_0 = (\mathcal{E}_\mathrm{ge} \cap \mathcal{E}^\ast) \cup (\mathcal{E}_\mathrm{as} \cap \mathcal{E}^\ast)
\end{equation}

We encode the discrete concept tokens with the CLIP text encoder then perform $L_2$ normalization to obtain the base and candidate concept embedding matrices:
\begin{equation}
    W_0 = \Phi_T(\mathcal{E}_0),\quad W^\ast = \Phi_T(\mathcal{E}^\ast)
\end{equation}

\begin{figure}[t]
    \centering
    \includegraphics[width=\linewidth]{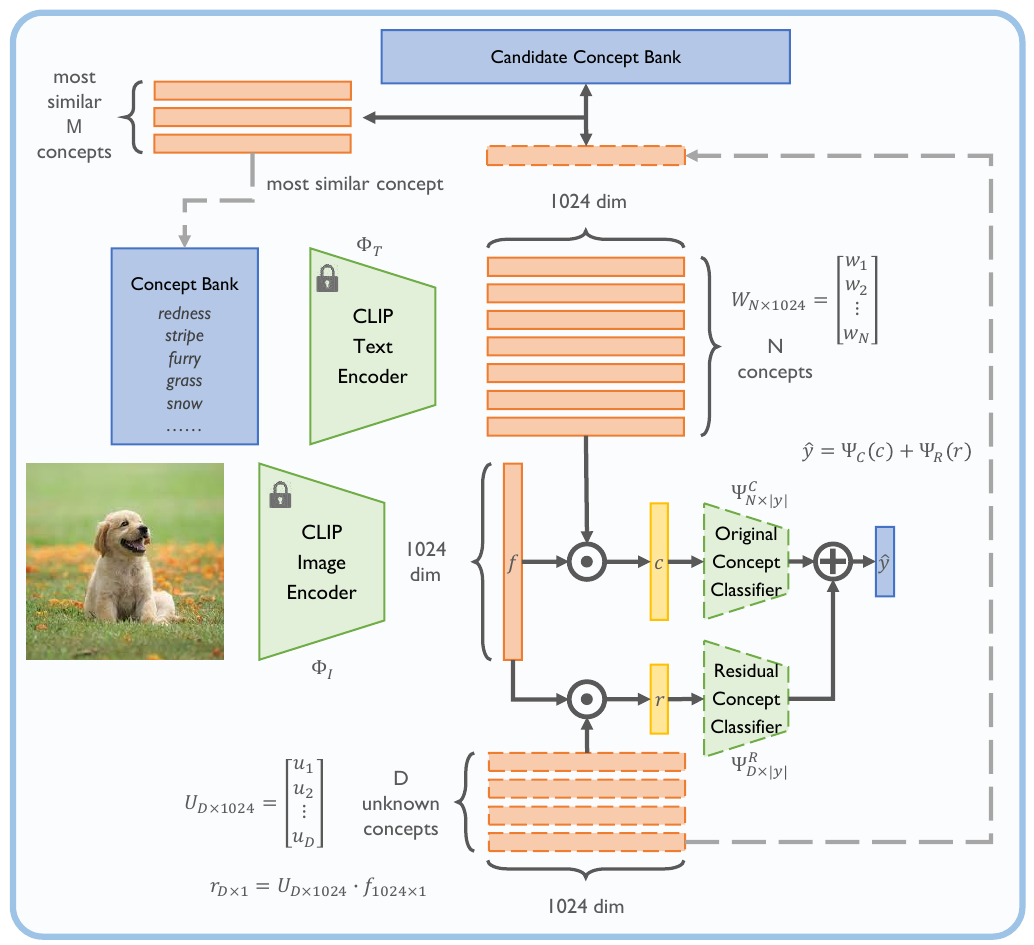}
    \caption{Incremental Residual Concept Bottleneck Model.}
    \label{fig:rescbm}
\end{figure}


\subsection{Residual Concept Bottleneck Model}
\label{subsec:rescbm}

Previous methods have typically involved pre-generating a set of relevant concepts and then deleting some of them to get the final concept bank. However, on the one hand, the pre-generated concepts might be insufficient, as their scale has often limitations. On the other hand, manual filtering without leveraging dataset information may result in the omission of important concepts. Therefore, we propose the residual concept bottleneck model to learn the missing concepts from the dataset content in an incremental approach compared to other decremental methods as shown in Fig.\ref{fig:rescbm}. 

We first randomly initialize $D$ optimizable vectors $U_{D \times d}=\{u_1,u_2,...,u_D\}$ with dimensions consistent with those of the existing concept embeddings. We likewise project the visual representations onto these vectors, resulting in residual concepts $r = U \cdot f$ with unclear meanings. The prediction result of using the residual concepts to complement the original concepts is as follows:
\begin{equation}
    \hat{y} = \Psi_c(c) + \Psi_r(r) = \Psi_c(W \cdot f) + \Psi_r(U \cdot f)
\end{equation}
where $\Psi_c:\mathbb{R}^N \rightarrow \mathcal{Y}$ is the original concept linear classifier using existing concepts and $\Psi_r:\mathbb{R}^D \rightarrow \mathcal{Y}$ is the residual concept linear classifier using unknown residual concepts.

For one thing, we intend to maximize the accuracy when using only the original concepts, which can be formulated as the following optimization problem:
\begin{equation}
    \min_{\Psi_c} \mathop{\mathbb{E}}\limits_{(x,y) \in \mathcal{D}} \Big[ \mathcal{L} \big( \Psi_c(W \cdot \Phi_I(x)),y \big) \Big] + \lambda \cdot \Omega(\Psi_c)
\end{equation}
where $\mathcal{L}(\hat{y},y)$ is the cross-entropy loss function, $\Omega(\Psi_c)$ is a complexity measure to regularize the model, and $\lambda$ is the regularization strength.

From another perspective, we aim to improve the performance after incorporating the residual concepts, which can be solved by the following optimization problem:
\begin{equation}
    \min_{\Psi_r,U} \mathop{\mathbb{E}}\limits_{(x,y) \in \mathcal{D}} \Big[ \mathcal{L} \big( \Psi_c(W \cdot f) + \Psi_r(U \cdot f),y \big) \Big] + \lambda \cdot \Omega(\Psi_r)
\end{equation}
where $f = \Phi_I(x)$.

To simultaneously address these two problems, we perform two forward and backward propagations in each batch to update the parameters. When performing the first propagation, we utilize only the original concepts to optimize the original concept classifier $\Psi_c$. During the second propagation, we fix the parameters of the original concept classifier $\Psi_c$ and optimize the residual concept classifier $\Psi_r$ as well as the learnable unknown concept vectors $U$ using both the original and the residual concepts.

To avoid the influence of extreme values from unknown residual concepts, we employ cosine similarity instead of projection distance as the concept bottleneck and standardize the activations of each concept to have a mean of $0$ and a standard deviation of $1$ to speed up convergence \cite{oikarinen2023label}. These operations are reversible and do not affect the relative order of concepts or introduce additional information.


\subsection{Incremental Concept Discovery}
\label{subsec:icd}

Through the aforementioned steps, the residual concepts effectively compensate for the incompleteness of the original concepts. However, these randomly learned concepts lack human interpretability, so we next translate them into human-understandable concepts.

\textbf{Parameters Initialization.} We first initialize another optimizable vector $v_d$, denoted as the discovered concept vector, whose value is the mean concept embeddings of the base concept bank with added noise. Subsequently, we append the discovered concept vector to the base concept bank and remove one of the learned unknown residual concept vectors, whose residual concept embedding matrix changes to $U_{-v_d} = U_{(D-1) \times d}$ and residual classifier changes to $\Psi_{r-v_d}:\mathbb{R}^{D-1} \rightarrow \mathcal{Y}$. After performing $L_2$ normalization and projection, the visual representations will also obtain a similarity score $c_d = v_d \cdot f$ to the discovered concept vector, and there will be a corresponding discovered concept linear classifier $\Phi_d:\mathbb{R} \rightarrow \mathcal{Y}$. We adopt the language priors as the initial weights to ensure stable updating of the discovered concept vector. In particular, the weights of the discovered concept classifier are initialized as the cosine similarity between the discovered concept embedding and the text embeddings for each class name, while other parameters and weights are kept untouched.

\textbf{Concept Similarity Loss Function.} After adding the discovered concept vector to the base concept bank, the prediction under the original and residual strategy is as follows:
\begin{gather}
    \hat{y}_{\mathrm{org}} = \Psi_c(c) + \Psi_d(c_d) = \Psi_c(W \cdot f) + \Psi_d(v_d \cdot f) \\
    \hat{y}_{\mathrm{res}} = \hat{y}_{\mathrm{org}} + \Psi_{r-v_d}(U_{-v_d} \cdot f)
\end{gather}

To further guarantee that the discovered concept vector is interpretable, we propose the concept similarity loss function $\mathcal{L}_\mathrm{sim}$, which utilizes the candidate concept bank $W^*$ to constrain the discovered concept embedding:
\begin{equation}
    \mathcal{L}_\mathrm{sim} = 1 - \mathbb{E} \Big[ \mathrm{Rank}\big(W^* \cdot \frac{v_d^T}{\Vert v_d \Vert_2}, M \big) \Big]
\end{equation}
where $\mathrm{Rank}(v,M)$ denotes sorting $v$ from largest to smallest and selecting the first $M$ of them.

Referring to the update strategy of the residual concept bottleneck model, we just add the discovered concept with its classifier during the first pass propagation:
\begin{equation}
    \min_{\Psi_c,\Psi_d,v_d} \mathbb{E} \big[ \mathcal{L} ( \hat{y}_\mathrm{org},y ) \big] + \alpha \cdot \mathcal{L}_\mathrm{sim} + \lambda \cdot \Omega(\Psi_c,\Psi_d)
\end{equation}
where $\alpha$ is the weight of the concept similarity loss.

When performing the second propagation, we fix the parameters of the original concept classifier $\Psi_c$, the discovered concept classifier $\Psi_d$, and the discovered concept embedding $v_d$, optimizing the residual concept classifier $\Psi_{r-v_d}$ and the unknown residual concept vectors $U_{-v_d}$:
\begin{equation}
    \min_{\Psi_{r-v_d},U_{-v_d}} \mathbb{E} \big[ \mathcal{L} ( \hat{y}_\mathrm{res},y ) \big] + \lambda \cdot \Omega(\Psi_{r-v_d})
\end{equation}

We recover concepts in an incremental way until all the learned unknown residual concept vectors are transformed into pure, accurate, and human-interpretable concepts. At this point, the base concept library is fully updated,  and experiments confirm its superior completeness.
\section{Experiments}
\label{sec:experiments}

\subsection{Experimental Setup}
\label{subsec:setup}

\textbf{Datasets.} We conduct comprehensive experiments on 7 datasets, including: CIFAR-10, CIFAR-100 \cite{krizhevsky2009learning}, Tiny-ImageNet \cite{le2015tiny}, LAD \cite{zhao2019large}, CUB-200 \cite{wah2011caltech}, Food-101 \cite{bossard2014food}, Flower-102 \cite{nilsback2008automated}. For a fair comparison, we apply CLIP-RN50 when comparing with other CBMs and CLIP-ViT-L/14 on few-shot classification tasks as the backbone, and all linear classifiers are optimized with the Adam \cite{kingma2014adam} optimizer. For more detailed information, please refer to the supplementary material. 

\textbf{Baselines.}
To evaluate our proposed approach, we compare it with the following baseline methods: 


\begin{table*}[t]
\centering
\footnotesize
\caption{Comparison with other CLIP-based CBMs. Bold indicates the best interpretable result, underline indicates the 2nd-best result, italic indicates the black-box methods.}
\begin{tabularx}{\linewidth}{*{9}{>{\centering\arraybackslash\hsize=0.95\hsize}X}}
\toprule
\multirow{2}{*}{Method} & \multirow{2}{*}{Interpretability} & \multirow{2}{*}{Length} & \multicolumn{3}{c}{CIFAR 10} & \multicolumn{3}{c}{CIFAR 100} \\
\cmidrule(lr){4-6} \cmidrule(lr){7-9}
& & & Accuracy & Number & CUE & Accuracy & Number & CUE \\
\midrule
Zero-shot & Yes & N/A & 0.6796 & N/A & N/A & 0.2620 & N/A & N/A \\
LP & No & N/A & 0.8858 & N/A & N/A & 0.6995 & N/A & N/A \\
PCBM-1r & Yes & 9 & 0.8044 & 175 & \textbf{5.1073} & 0.5379 & 440 & 1.3583 \\
PCBM-2r & Yes & 9 & 0.8074 & 383 & 2.3423 & 0.5450 & 1175 & 0.5153 \\
PCBM-3r & Yes & 9 & 0.8208 & 1058 & 0.8620 & N/A & N/A & N/A \\
PCBM-h & No & 9 & 0.8757 & 175 & N/A & 0.6683 & 440 & N/A \\
Lf-CBM & Yes & 12 & 0.8677 & 143 & 5.0565 & {\ul 0.6745} & 892 & {\ul 0.6301} \\
LaBo-10c & Yes & 27 & 0.8489 & 100 & 3.1441 & 0.6618 & 1000 & 0.2451 \\
LaBo-20c & Yes & 27 & 0.8669 & 200 & 1.6054 & 0.6736 & 2000 & 0.1247 \\
LaBo-30c & Yes & 26 & {\ul 0.8752} & 300 & 1.1221 & N/A & N/A & N/A \\
\textbf{Res-CBM} & Yes & 7 & \textbf{0.8803} & 237+10 & {\ul 5.0914} & \textbf{0.6791} & 372+15 & \textbf{2.5396} \\
\bottomrule
\end{tabularx}
\label{tb:clip-cbms}
\end{table*}


\begin{table*}[h]
\centering
\footnotesize
\caption{Comparison with CBMs with labeled concept annotations. The labeling significance is consistent with Tab.\ref{tb:clip-cbms}.}
\setlength{\tabcolsep}{12pt}
\begin{tabular}{>{\centering\arraybackslash}m{2.5cm}*{7}{c}}
\toprule
Method & LAD-A  & LAD-E  & LAD-F  & LAD-H  & LAD-V  & CUB & \textbf{Mean}  \\
\midrule
LP & 0.9432 & 0.8661 & 0.7393 & 0.4924 & 0.8671 & 0.7214 & 0.7714 \\
Original CBM & {\ul 0.8964} & {\ul 0.7826} & 0.5535 & {\ul 0.3388} & {\ul 0.7907} & \textbf{0.6513} & {\ul 0.6689} \\
PCBM-annotation & 0.8934 & 0.7711 & {\ul 0.5982} & 0.2982 & 0.7814 & 0.6363 & 0.6631 \\
Res-CBM-before & 0.7897 & 0.7709 & 0.4356 & 0.1429 & 0.7548 & 0.5987 & 0.5812 \\
\textbf{Res-CBM-after} & \textbf{0.8979} & \textbf{0.8045} & \textbf{0.6624} & \textbf{0.3862} & \textbf{0.8301} & {\ul 0.6243} & \textbf{0.7009} \\
\bottomrule
\end{tabular}
\label{tb:labeled-cbms}
\end{table*}


\begin{itemize}

\item{CLIP Zero-shot}\cite{radford2021learning}: the cosine similarities between the visual representations and the textual embeddings of each class name are calculated directly without any training. The class with the highest cosine similarity score is selected as the final classification result. 

\item{Linear Probing (LP)}\cite{radford2021learning}: We exclusively utilize the CLIP visual encoder and the obtained visual representations are then used to predict the label by training a linear classifier. 

\item{Original CBM}\cite{koh2020concept}: The original CBM relies on fine-grained concept annotations, so we compare our method with it only on LAD and CUB with available attribute annotations. In addition, CBMs may suffer from concept leakage, which can only be alleviated by CBMs trained independently \cite{margeloiu2021concept}. Hence, we choose CBM-independent as the baseline for comparison.

\item{PCBM}\cite{yuksekgonul2022post}: PCBM uses the ConceptNet\cite{speer2017conceptnet}, a knowledge graph dataset, to conduct a concept bank. It treats classes as nodes and incorporates surrounding neighboring nodes in the concept bank. More concepts can be included by expanding multiple layers outwards.

\item{PCBM-h}\cite{yuksekgonul2022post}: After training the PCBM, PCBM-h is applied as a post-processing method that connects the visual representation and the class labels result through shortcuts. It utilizes a linear classifier to predict the difference between the result of PCBM and the ground truth.

\item{Lf-CBM}\cite{oikarinen2023label}: In contrast to PCBM, Lf-CBM generates concepts through a large language model (LLM). It solely utilizes CLIP as the concept annotator, allowing it to act on any visual backbone. We directly select CLIP as the visual backbone, thus omitting the process of computing the similarity matrix.

\item{LaBo}\cite{yang2023language}: LaBo similarly applies the LLM to generate candidate concepts. However, the quantity of concepts in LaBo is specific to each class. So we generate 10, 20, and 30 concepts per class to compare the performance. In addition, LaBo's linear classifier differs from the above baselines. To evaluate the quality of concepts equitably, we choose the same classifier as the above methods.
\end{itemize}

\textbf{Metrics.}
Besides accuracy, we also concentrate on concepts' efficiency, including both the average length of each concept and quantity of concepts. Specifically, we define the accuracy improvement brought by all concepts' length to measure the \textbf{C}oncept \textbf{U}tilization \textbf{E}fficiency ($\mathrm{CUE}$):
\begin{equation}
    \mathrm{CUE} = \frac{10000\times \mathrm{Acc}}{N \times \bar{L}}
\end{equation}
where $\mathrm{Acc}$ is the classification accuracy, $N$ is the quantity of concepts and $\bar{L}$ is the average number of letters included in each concept. Larger $\mathrm{CUE}$ means higher efficiency.


\subsection{Accuracy and Efficiency}
\label{subsec:accuracy}

We compare our method with other CBMs in terms of accuracy and efficiency. Tab.\ref{tb:clip-cbms} provides the results of our method compared to other CLIP-based CBMs.\footnote{PCBM-\#r represents expanding outward to include neighboring nodes \# times. LaBo-\#c represents selecting \# concepts per class.} Res-CBM attains higher accuracy than PCBM by 5.95\% on CIFAR-10 and 13.41\% on CIFAR-100 and even outperforms the inexplicable PCBM-h. Despite Lf-CBM and LaBo being powerful baselines, our methodology still achieves state-of-the-art results. Particularly, it should be noted that our method considerably exceeds both Lf-CBM and LaBo in terms of efficiency, as evidenced by the $\mathrm{CUE}$ metric.


\begin{figure*}[t]
    \centering
    \includegraphics[width=0.95\linewidth]{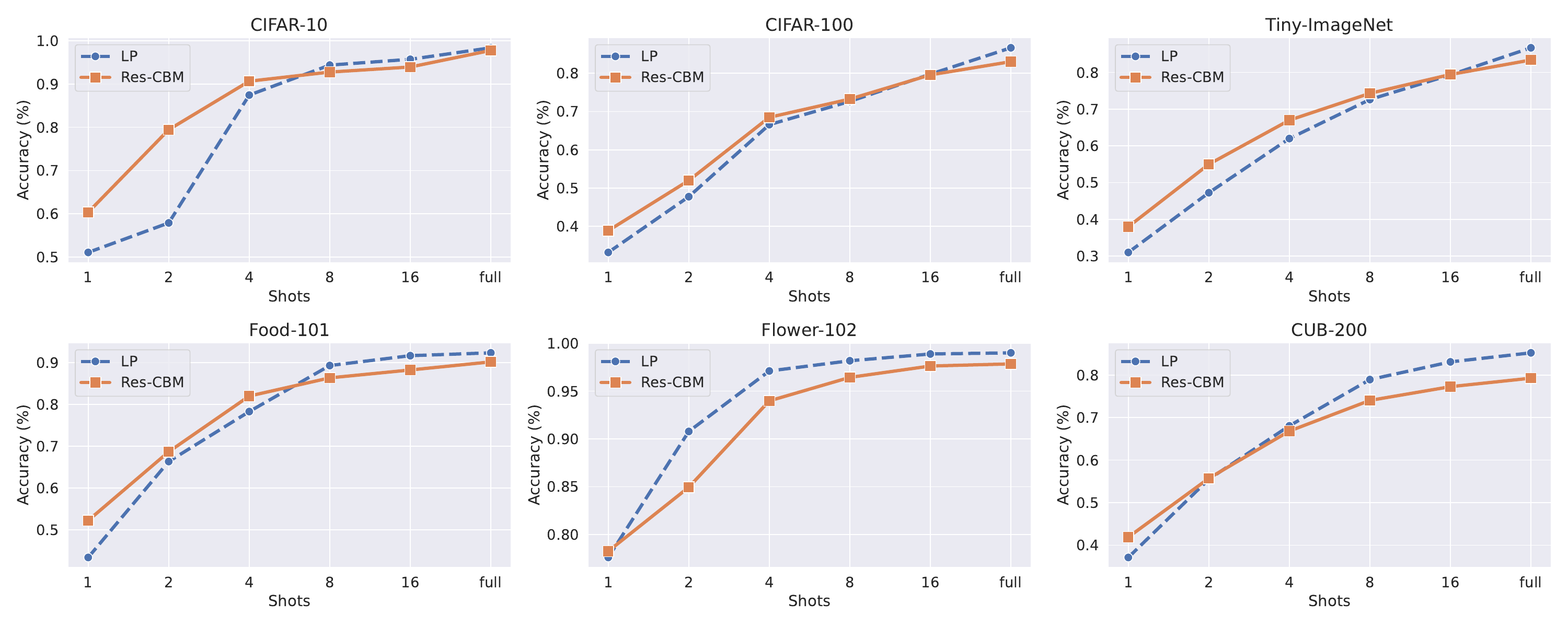}
    \caption{Test accuracy (\%) comparison between Res-CBM and LP on 6 datasets. The x-axis represents the number of labeled images.}
    \label{fig:vs-lp}
\end{figure*}

\begin{figure*}[t]
    \centering
    \includegraphics[width=0.95\linewidth]{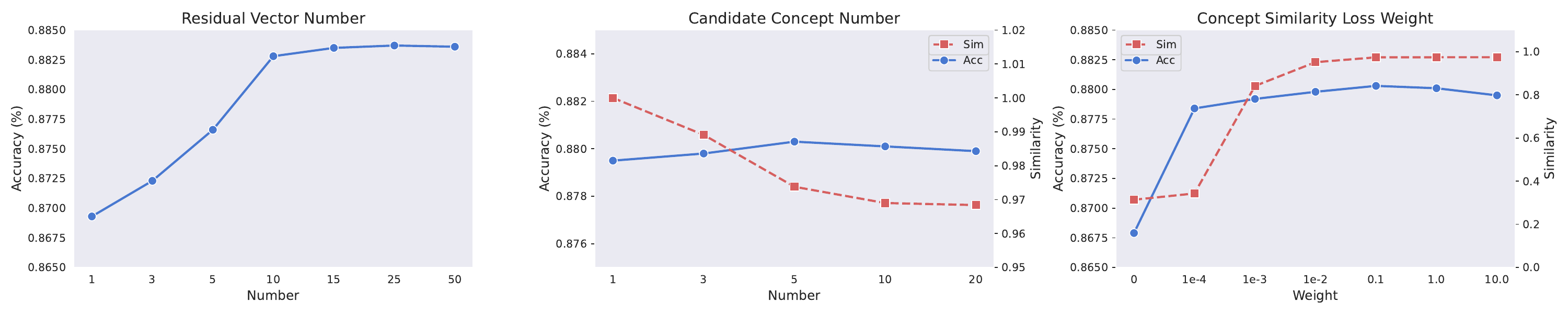}
    \caption{Ablation results on residual vector numbers, concept similarity loss weights and candidate concept numbers.}
    \label{fig:ablation}
\end{figure*}


On LAD and CUB-200 with attribute annotations, we still use CLIP-50 with frozen parameters as the visual backbone to train Original CBM according to formula (\ref{eq:cbm}), as the results are shown in Tab.\ref{tb:labeled-cbms}.\footnote{PCBM-annotation refers to PCBM that utilizes the textual embeddings of the annotated attributes as the concept bank. Res-CBM-before indicates the performance on the initial base concept bank. Res-CBM-after represents the performance after incremental concept discovery.} The close performance of the Original CBM and PCBM-annotation suggests that the CBMs' performance heavily relies on the expert-selected concepts more than the model structure. Although Res-CBM-before performs poorly, a remarkable improvement is observed in Res-CBM-after after applying our unsupervised incremental concept discovery, which even surpasses supervised CBMs with concept annotations on most datasets.

\subsection{Interpretability}
\label{subsec:interpretability}

\begin{figure}[t]
    \centering
    \includegraphics[width=0.9\linewidth]{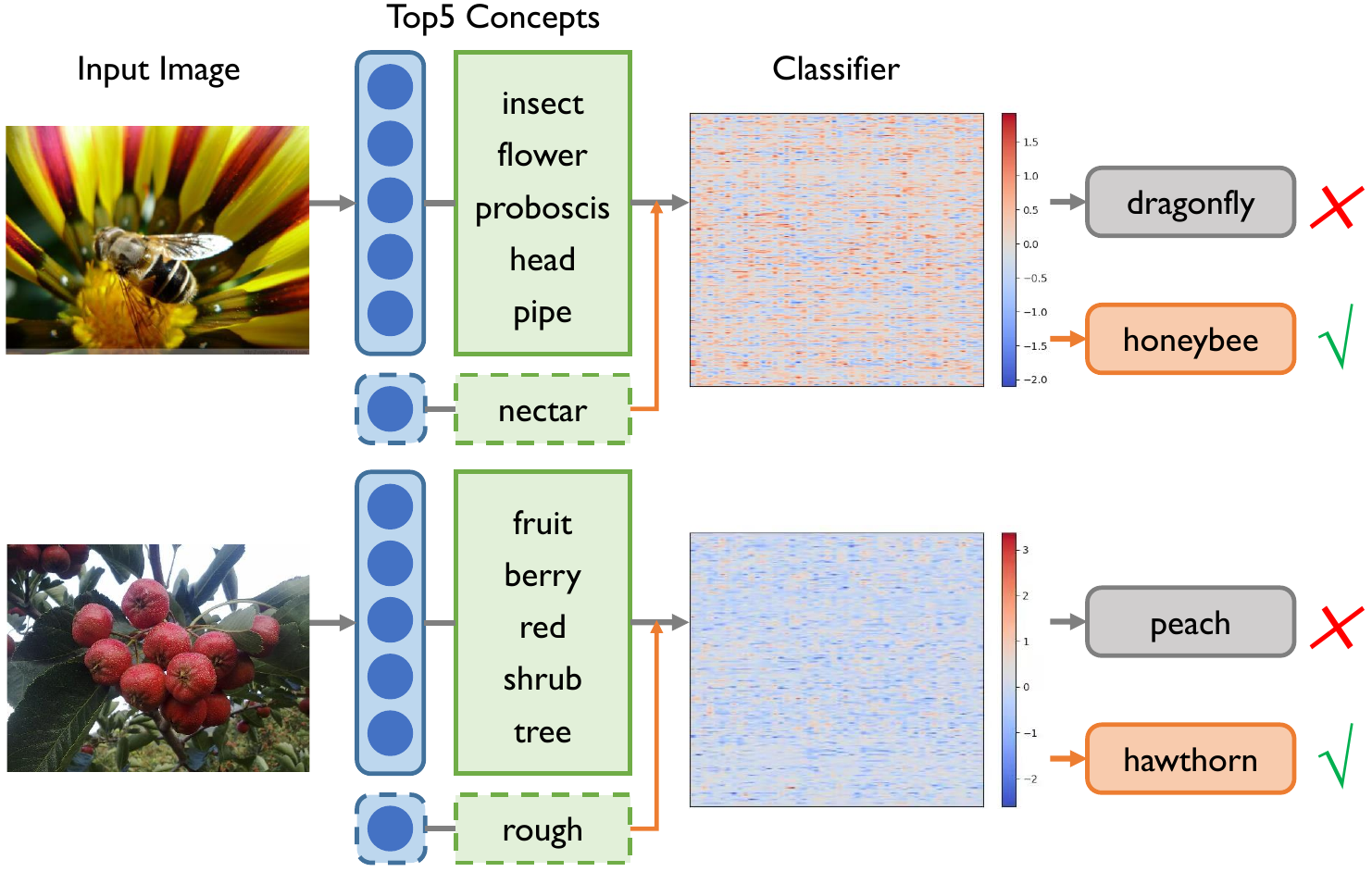}
    \caption{Concepts applied in Res-CBM on LAD-A (top) and LAD-F (bottom). Solid boxes represent concepts from the base concept bank, dashed boxes indicate the discovered concepts. Heatmaps refer to learned linear concept classifiers, in which the x-axis represents classes, the y-axis represents concepts, and the values represent the contribution of concepts to categories.}
    \label{fig:intepret}
\end{figure}

In this subsection, we visualize the mechanism of Res-CBM and demonstrate how the newly discovered concepts assist in advancing classification performance. For each image, we display the top 5 concepts with the highest activation scores in the concept bottleneck, as shown in Fig.\ref{fig:intepret}. Through a sparse concept classifier, we can obtain predictions, which signifies predicting the class by a linear combination of concepts.

However, due to the incompleteness of the base concept bank, it is challenging to achieve accurate classifications by solely relying on these concepts in several images, e.g., \textit{dragonfly} and \textit{honeybee}, \textit{peach} and \textit{hawthorn}, both of which possess concepts from the base concept bank. With the help of Res-CBM, we have discovered 2 concepts, \textit{nectar} and \textit{rough}, from the candidate concept bank. Unlike dragonflies, honeybees often gather nectar from the center of flowers, thus associating them with the concept of nectar. Moreover, hawthorns have a rougher surface compared to peaches. By incorporating these discovered concepts into the base concept bank, these initially misclassified samples are now correctly categorized, achieving automatic model debugging to a certain extent.


\subsection{Few-shot Capability}
\label{subsec:fewshot}

Fig.\ref{fig:vs-lp} illustrates the performance comparison under different data volume settings between Res-CBM and LP. Compared to the end-to-end black-box model, our method achieves superior performance when little data is available, and exhibits a slight performance gap with larger amounts of sample sizes, which indicates that our method has maintained the accuracy without sacrificing interpretability.

For common object datasets, our approach demonstrates excellent performance with 1, 2, 4, and 8 labeled images per category. We credit its success to the incorporation of human concept information, which helps the model extract vital aspects relevant to the categories from the representations of the black-box backbone.

For fine-grained object datasets, our method outperforms the black-box model when the data volume is extremely low. As the data volume increases, the unexplainable black-box model's performance becomes more prominent, especially in expert datasets such as Flower-102 and CUB-200. It might be attributed to common concepts struggling to distinguish the subtle differences between these fine-grained objects. A feasible solution is to enhance the specificity of concepts by a tailored specialized concept candidate bank.

\subsection{Ablation Study}
\label{subsec:ablation}


\textbf{Residual Vector Number.} To investigate the impact of residual vector numbers on performance improvement and demonstrate that our method can be applied as a post-processing approach to any CBMs, we record the results of attaching different residual vector numbers for Lf-CBM on CIFAR-10, as shown in Fig.\ref{fig:ablation}-left. As the quantity of vectors increases, the performance of Lf-CBM continuously improves and gradually saturates at around 15 vectors.

\textbf{Candidate Concept Number.} We fix the residual vector number at 10 and set the concept similarity loss weight to 0.1, then record the final accuracy and the average similarity between the discovered concepts and the most similar concepts in the candidate concept bank during the discovery process in Fig.\ref{fig:ablation}-middle. When the candidate concept number is small, the model tends to converge to this unique concept, making it difficult to escape local optima. When it is large, the optimizing vectors easily oscillate between different concepts, resulting in lower accuracy and similarity. Overall, setting the candidate concept number to 5 seems to be a good hyperparameter choice.

\textbf{Concept Similarity Loss Weight.} With a fixed candidate concept number at 5, we also record the final accuracy and average similarity for different loss weights as shown in Fig.\ref{fig:ablation}-right. When the weight is low, the similarity of the discovered concepts is extremely low, which resembles a random selection strategy, leading to poorer performance. When the weight is large, the model tends to converge to a local optimum only satisfing the concept similarity loss, sacrificing the ability to explore more appropriate concepts. Since Res-CBM has already saturated after the completion of unknown concept vectors, the cross-entropy loss value is small. Therefore, a weight of 0.1 realizes optimal results.
\section{Discussion and Conclusion}
\label{sec:conclusion}

\textbf{Rethinking CBMs.}
Indeed CBMs are valuable attempts towards transparency of the decision-making process, one important and fundamental problem may be still unclear, which is: how many concepts are good? especially when lacking a well-accepted and rigorous definition of concepts. In this paper, we have tried to explore this problem. 
Firstly, the number of concepts should not be the larger the better. From the optimization perspective, the operation of CBMs to project visual representations into concept space is a linear transformation. If its rank is greater than or equal to the dimension $d$ of the concept vector, this process is invertible. Therefore, it is straightforward to re-estimate the inverse transformation matrix to preserve classification accuracy in subsequent linear classifiers. The number of concepts larger than $d$ may decrease the efficiency, making the interpretability less effective.

On the other hand, the concept should be an efficient way to express categories. Humans can describe objects in the world exponentially using a finite set of concepts. Theoretically, from informatics, when each concept weight is binary, the concept quantity needed to distinguish $n$ categories will be enough approximately $\log_2{n}$. This is the reason for the poor $\mathrm{CUE}$ of LaBo \cite{yang2023language}, which finds concepts for each category in a way that the $\mathrm{CUE}$ complexity is $\mathcal{O}(n)$ rather than $\mathcal{O}(\log_2{n})$. This suggests that the $\mathrm{CUE}$ is crucial in CBMs, and a reasonable number of concepts should lie in the interval of $\big[\lceil\log_2{n}\rceil,d\big)$.

\textbf{Conclusion.} In this paper, we emphasize the significance of purity, precision, and completeness in CBMs and introduce the $\mathrm{CUE}$ metric to evaluate the descriptive efficiency of CBMs. Our proposed Res-CBM complements the missing concepts in any CBM with an incremental concept discovery approach, which achieves optimal accuracy with the best concept utilization efficiency. Experiments demonstrate that our approach outperforms end-to-end black-box models in few-shot learning and surpasses other CLIP-based CBMs even the original CBM with supervised concept annotations on most datasets.

\textbf{Limitation and Future Work.} For our proposed Res-CBM, the candidate concept bank construction for fine-grained datasets remains a challenge. Additionally, the incremental and sequential concept discovery approach increases computational time costs. In the future, we will develop more efficient concept similarity calculation and explore parallel concept discovery techniques. Regarding the CBM structure, existing CBMs represent all concepts into a single bottleneck, disregarding the hierarchical nature of concepts, where simple concepts can be combined to form more complicated ones. Building efficient CBMs that align with human cognitive processes will contribute to more transparent artificial intelligence systems.

\section*{Acknowledgements}
This work was partly supported by the National Natural Science Foundation of China (Grant No. 61991451) and the Shenzhen Science and Technology Program (JCYJ20220818101001004). The first two authors have equal contribution.

{
    \small
    \bibliographystyle{ieeenat_fullname}
    \bibliography{main}
}

\end{document}